\documentclass[a4paper, 11pt]{article}
\usepackage[english]{babel}
\usepackage[utf8]{inputenc} 
\usepackage[T1]{fontenc}    
\usepackage{hyperref}       
\usepackage{url}            
\usepackage{amsfonts}       
\usepackage{nicefrac}       
\usepackage{amsmath}
\usepackage{bbm}
\usepackage{blkarray}
\usepackage{graphicx}
\usepackage{subfigure}
\usepackage{color}
\usepackage{multirow}
\usepackage{fullpage}
\usepackage{amsthm}
\usepackage{amssymb}

\newcommand{\arxiv}[1]{#1}
\newcommand{\conference}[1]{}
\makeatletter
\newcommand{\printfnsymbol}[1]{%
  \textsuperscript{\@fnsymbol{#1}}%
}
\makeatother

\title{WEST: Word Encoded Sequence Transducers}
\author{Ehsan Variani\thanks{equal contribution}, Ananda Theertha Suresh\footnotemark[1], Mitchel Weintraub \\
\texttt{$\{$variani, theertha, mweintraub$\}$@google.com} \vspace{3ex} \\
Google Inc.
}

\begin{document}

\maketitle
  \newcommand{\tdef}{\stackrel{\text{def}}{=}}
  \newcommand{\ignore}[1]{}

\begin{abstract}
\conference{\vspace{0.1cm}}
 Most of the parameters in large vocabulary models are used in
 embedding layer to map categorical features to vectors and in softmax layer for classification weights.
 This is a bottle-neck in memory constraint on-device training applications
 like federated learning and on-device inference applications like automatic
 speech recognition (ASR). One way of compressing the embedding and softmax
 layers is to substitute larger units such as words with smaller sub-units such as characters.
 However, often the sub-unit models
 perform poorly compared to the larger unit models. We propose WEST, an algorithm
 for encoding categorical features and output classes with a sequence of random or domain dependent sub-units
 and demonstrate that this transduction can lead to significant compression
 without compromising performance. \arxiv{WEST bridges the gap between larger unit and sub-unit models and can be interpreted as a
 MaxEnt model over sub-unit features, which can be of independent interest.}
\end{abstract}

\section{Introduction}
\label{introduction}
\conference{A standard way of representing categorical features or classes as continuous features is to
use embedding look-ups e.g., word-to-vector representation
~\cite{mikolov2013efficient}.
The corresponding matrix which contains one embedding per row is called
embedding matrix and its size scales with the number of categorical features (or output classes) and
the embedding dimension. For the input layer (or \textit{embedding} layer), the
embedding matrix stores the representation for categorical features while for the
output layer (or \textit{softmax} layer), it stores the
classification weights corresponding to each output class.}
\arxiv{A standard way of converting categorical inputs to continuous features is to
  use embedding look-ups e.g., word-to-vector representation
  \cite{mikolov2013efficient}. If the number of categorical
features (referred to as vocabulary) is $V_{\text{in}}$ and the embedding
dimension is $d_e$, then storing all the embedding look-ups uses size
$V_{\text{in}} \times d_e$. The corresponding layer is called the input layer or
\textit{embedding} layer.
Embedding layers are used in a
wide variety of tasks including bag of words classification
models~\cite{zhang2010understanding}, language
models~\cite{mikolov2010recurrent,soltau2016neural}, and machine
translation~\cite{sutskever2014sequence}.

Similarly, for supervised learning tasks such as classification, the standard way of
training uses an output or \textit{softmax} layer, where each output class is associated
with a $d_s$ dimensional classification weight vector. Similar to embedding
look-ups, storing softmax vectors uses space
$V_{\text{out}} \times d_s$, where $V_{\text{out}}$ is the number of
output classes. Softmax layers are also commonly used for most
supervised learning tasks, including speech recognition, image recognition,
language models, and machine translation.}

For large vocabulary tasks\arxiv{ where $V_\text{in}$ and $V_\text{out}$ are large},
embedding \arxiv{ and softmax } layers might
not even fit within the memory capacity of a single accelerated computation
unit like Graphical Processing Unit (GPU)
\cite{kumar2017lattice, soltau2017reducing}. The situation is even worse when
using these models for on-device training such as federated learning
~\cite{konevcny2015federated,konevcny2016federated} or on-device inference
such as ASR on the phone \cite{mcgraw2016personalized} where there
is also communication bandwidth constraint between server, device and
CPU, accelerator respectively.
Furthermore, these models inherently suffer from unbalanced topology since
many core layers like recurrent layers are allocated a small percentage of the
total number of parameters; for Penn TreeBank (PTB) ~\cite{mikolov2010recurrent, zaremba2014recurrent}
or YouTube language model ~\cite{kumar2017lattice}, the embedding and softmax
layers contain nine times more parameters than the recurrent
layers\arxiv{ (Figure~\ref{fig:pie})}. Finally, due to data sparsity, there might not be enough training examples
per embedding parameter, particularly for the infrequent features. The
techniques to address the above challenges fall under a wide category of the
\textit{compression} algorithms.
\arxiv{
  \begin{figure}[t!]
  \centering
    \begin{tabular}{cccc}
 \includegraphics[width=.25\linewidth]{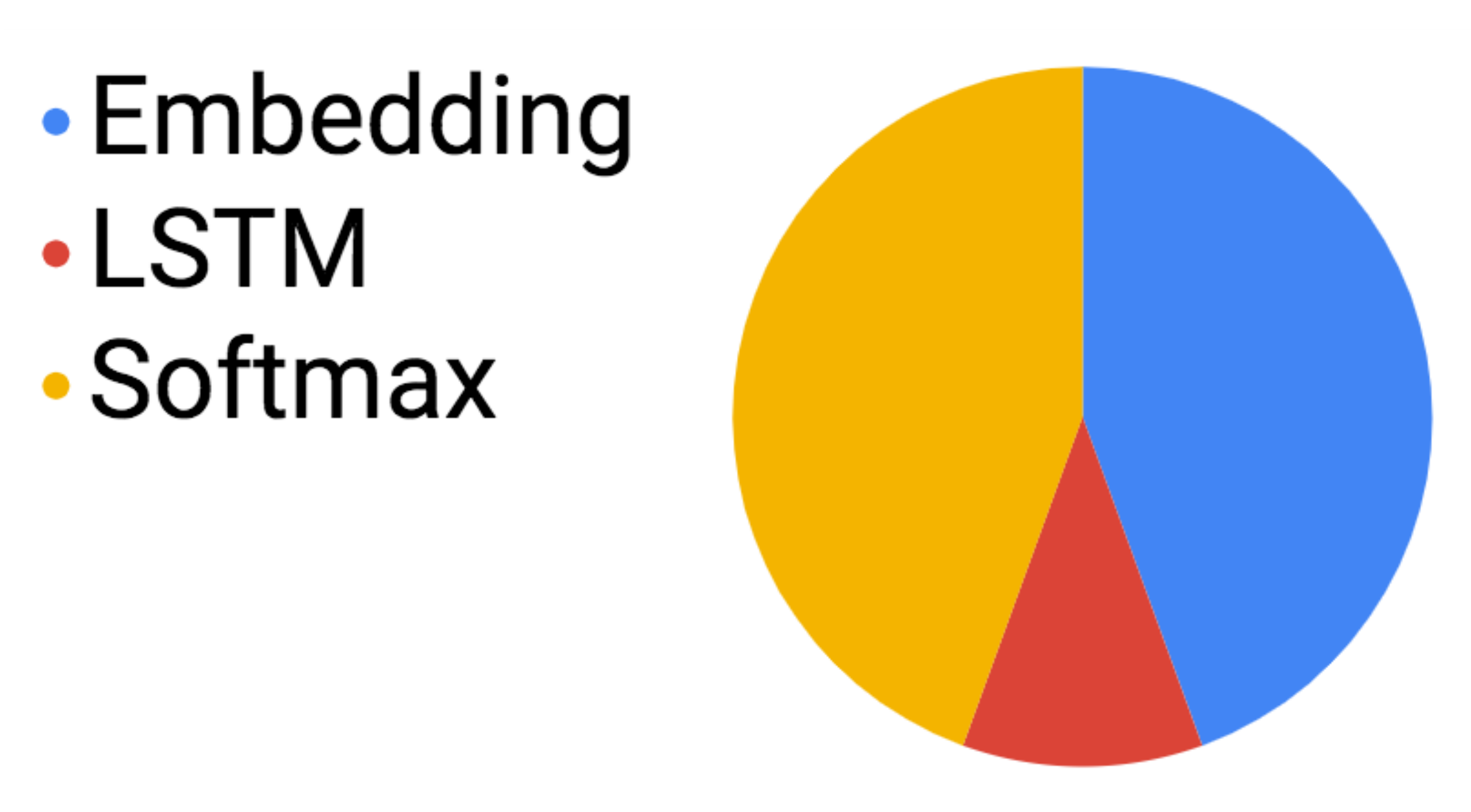} &
 \includegraphics[width=.16\linewidth]{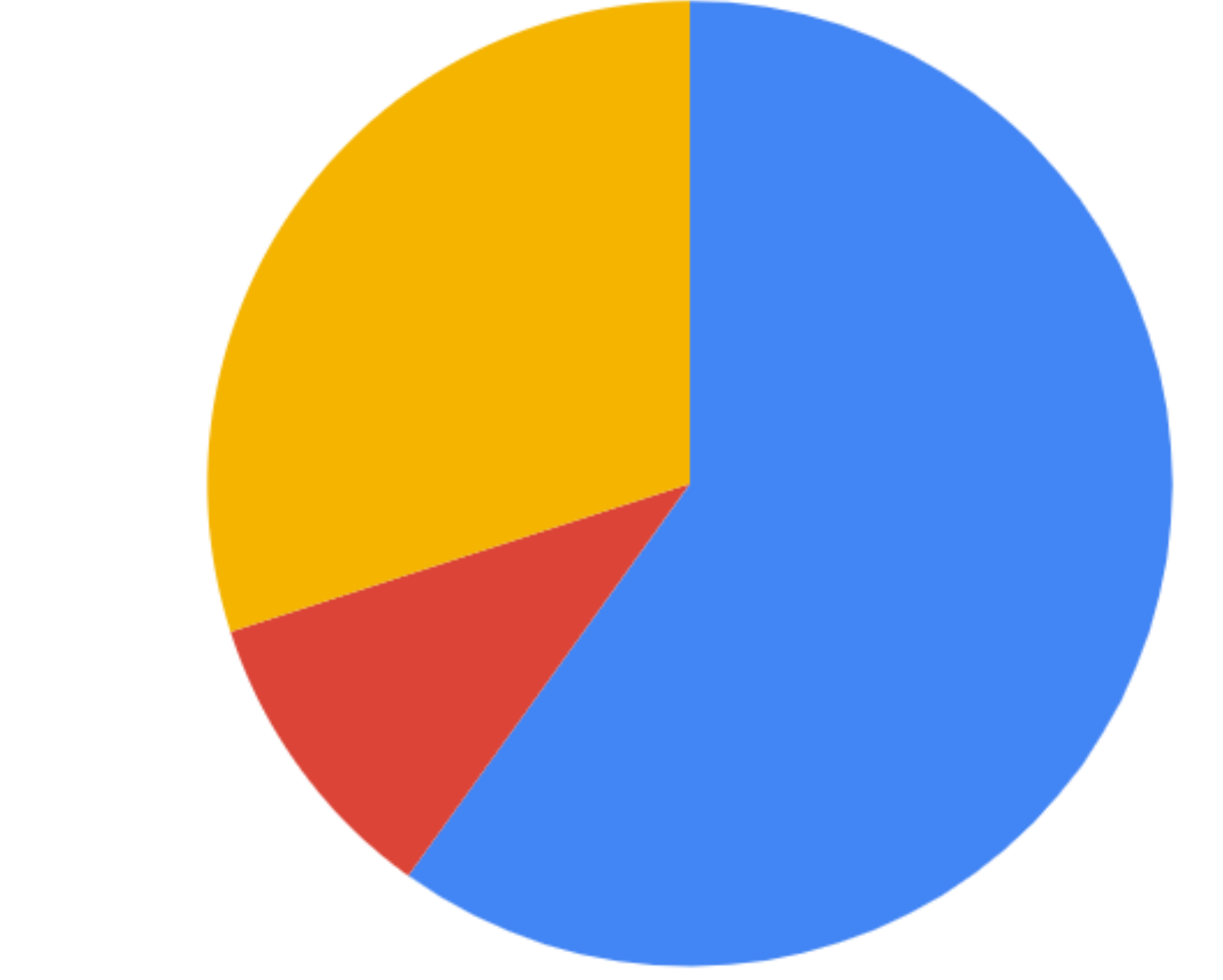} &
 \includegraphics[width=.18\linewidth]{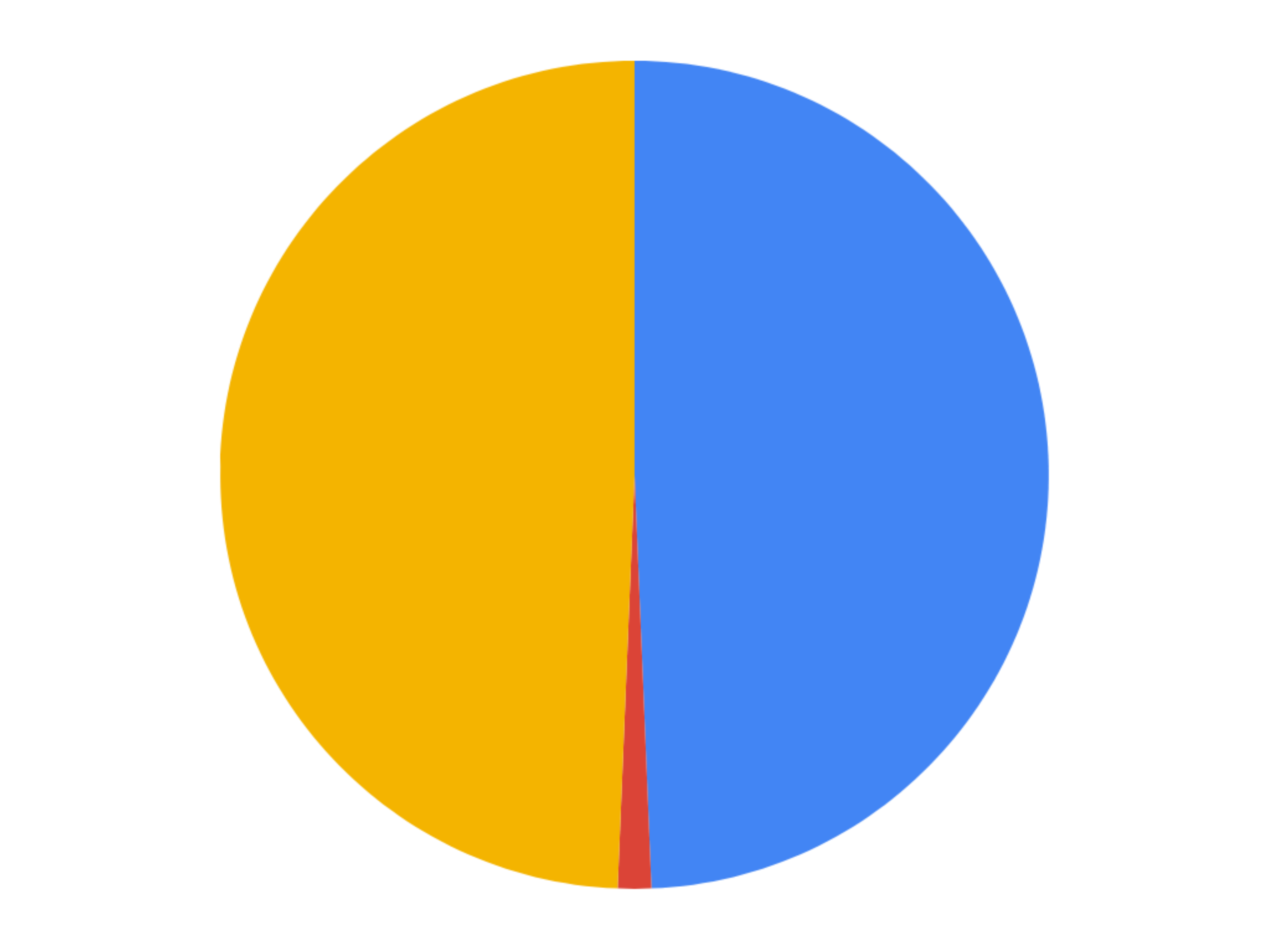}&
 \includegraphics[width=.18\linewidth]{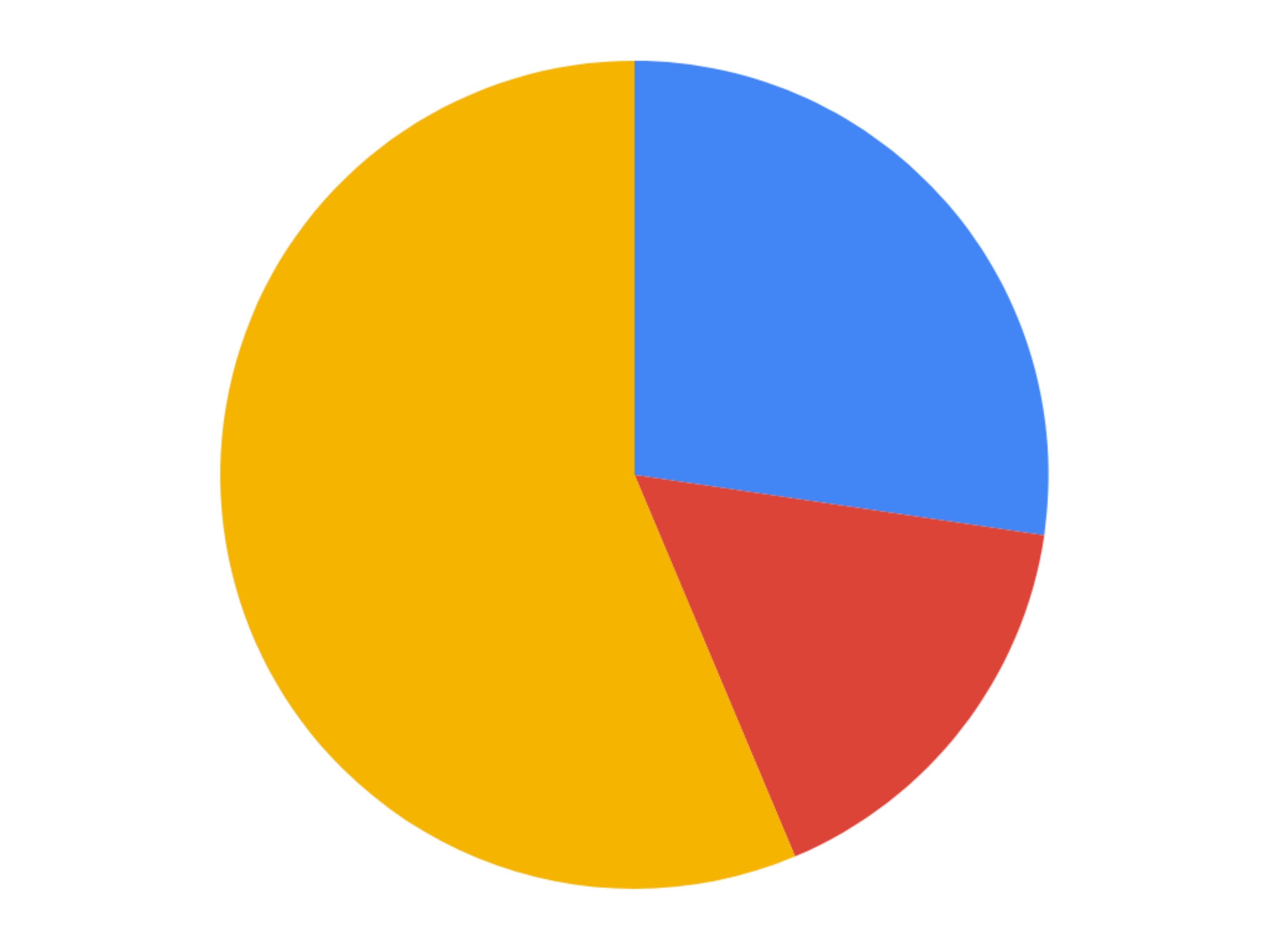}\\
 (a) Penn TreeBank~\cite{zaremba2014recurrent}. & YouTube \cite{kumar2017lattice}.
 & One-billion words~\cite{jozefowicz2016exploring}. & Embedded ASR (Sec.~\ref{sec:experiments}).
      \end{tabular}
  \caption{Model parameters for language models.}
    \label{fig:pie}
\end{figure}}

The compression algorithms are either \textit{multi-stage} or
\textit{single-stage}. In multi-stage compression techniques, a larger model
is first trained and then embedding or softmax layers are compressed via some
compression algorithm. This includes scalar
quantization~\cite{courbariaux2014training, anwar2015fixed,
  hwang2014fixed, wu2016google, alvarez2016efficient}, vector
quantization~\cite{gong2014compressing}, product
quantization~\cite{joulin2016fasttext}, combination of vector and
product quantization~\cite{kumar2017lattice}, Huffman
coding~\cite{han2015deep}, and low rank
approximation~\cite{chen2018groupreduce}.
These techniques do not optimize the same objective function as the neural
network training
objective function, thus often require another stage of retraining of the
uncompressed model parameters.\arxiv{ Furthermore, in applications such as federated
learning, they are infeasible as the model size needs to be small at
all steps of training.\par} Alternatively, single-stage approaches impose a
structure on the parameter space apriori of training and directly optimize the
smaller model. The most common
single stage compression technique for embedding layer is to use
hashing~\cite{chen2015compressing}, where the input vocabulary
is hashed to a smaller vocabulary. The hashing
mechanism is not one-to-one and is lossy.
Another common single-stage compression technique that also addresses data sparsity is to use sub-word units like word-pieces or characters. While these
methods naturally lead to high compression rate, there is still a gap between their
performance with that of larger unit models.

\arxiv{
Recently there are several papers that propose single stage
mechanisms~\cite{li2016lightrnn, li2017slim,
    chen2017learning, ling2015finding, shu2017compressing,
    chen2016compressing, verwimp2017character, oda2017neural}.
Of the above, perhaps the
closest to our work is Slim embedding~\cite{li2017slim},
which is a special case of WEST.
}
\conference{ This paper tries to shed light on the modeling performance gap between using smaller and larger units.}
\arxiv{
Our main contributions
are as follows.
\begin{itemize}
\item A general framework of structured sparse and structured dense
  decomposition for both embedding and softmax layers that
  generalizes~\cite{li2017slim}.
\item A MaxEnt intuition for the WEST softmax layer that bridges the gap between larger unit
  models and sub-units models.
\item Experimentally demonstrating that the resulting model performs
  well on variety of datasets and existing approaches.
  \end{itemize}
The rest of the paper is organized as follows. In
Section~\ref{sec:west}, we describe our method. In Section~\ref{sec:softmax}, we discuss the MaxEnt
interpretation for the softmax layer. In
Section~\ref{sec:experiments}, we discuss experiments on the
Penn TreeBank dataset and an on-device speech recognition system.
}

\conference{\vspace{-1ex}}\section{WEST: Word Encoded Sequence Transducers}
\label{sec:west}
\conference{\vspace{-1ex}}
\arxiv{
\begin{figure}[t]
  \centering
      \includegraphics[width=1.0\linewidth]{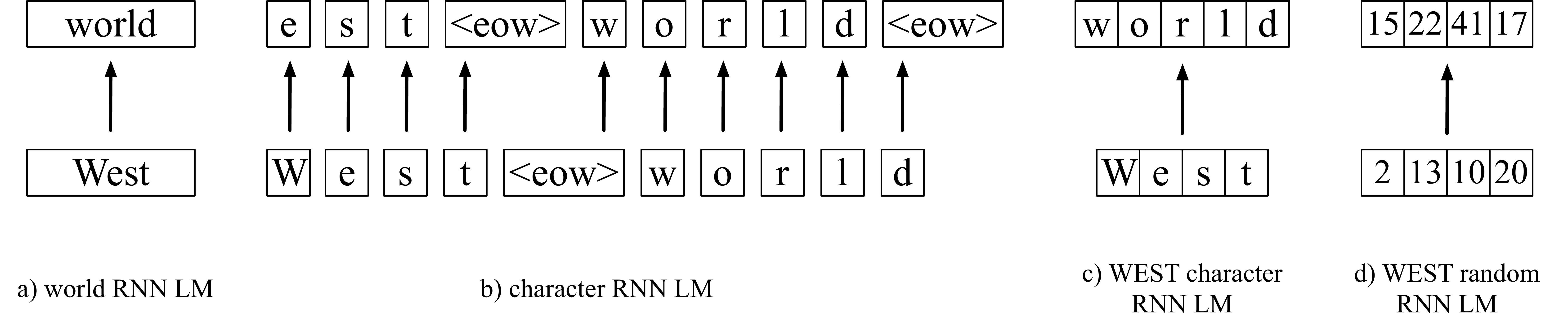}
      \caption{Different recurrent neural network language models (RNN LM)
       of sentence `West world'. (a) In the word model, the input word is fed to
      predict the next word. (b) In character model, the input and output operate on
      character level. (c, d) In WEST, while the word is represented by smaller
      sub-units (language dependent or random), the model still operates on larger units i.e.,
      word in this example.}
    \label{fig:rnn_lm}
\end{figure}
}
We hypothesize that the reason for
the performance difference between larger unit and smaller sub-unit models is
the way smaller unit models are structured.
Consider the example of language models in Figure~\ref{fig:rnn_lm}.
The main difference between word model and character model
is the cycle that these models operate on. In the word model, input word is fed
to predict the next word with word level cycle. The sub-unit models perform on
sub-unit cycle which is smaller than word.
Hence, the recurrent units need to remember longer sub-unit contexts.
Furthermore, since these models are normalized on sub-units, they
assign probabilities to character sequences that are not
in the vocabulary. Due to these reasons character models do not
perform as well as word models.

By using a model which performs over larger unit cycle
and encoding them with smaller units for internal representation, we show that the
performance gap gets smaller. In particular, we show that for the embedding, the larger
unit can be encoded internally using smaller sub-units and for the softmax, the output
class can be predicted by a MaxEnt model over sub-unit features. This allows us to
obtain space savings of smaller unit models, while keeping the model input and output
over larger units. We examine two different encodings, language and random encoding
and present experiments on a public data set as well as a large scale ASR task.
While the paper focuses on language model over words as it has both large embedding and
softmax layers, techniques can be applied to other models such as classification tasks
with ngram features. We now introduce the notion of encoding using sparse-dense matrix
factorization.

WEST is a single-stage compression technique that compresses the
embedding and softmax layers from the beginning of the training by
factoring them as a product of a structured sparse matrix and a
structured dense matrix.
Let $V$ be the vocabulary size i.e., the number of
categorical features or number of output classes and $d$ be the
embedding dimension. Then embedding and softmax matrices are of
size $V \times d$.
Let $E$ denote softmax or
embedding matrix.  We propose to factorize $E$ as
\[
E = C \times E^c,
\]
where $C$ is a sparse structured matrix of size $V \times n \cdot k$
such that each row of $C$ is concatenation of $n$ weighted one hot vectors of
length $k$
and $E^c$ is a structured dense matrix of size $n \cdot
k\times d$. \arxiv{If $C$ is the identity matrix and $E^c = E$, then
  this factorization is same as the normal embedding matrix.} Examples
of $C$, $E^c$\arxiv{, and $E$} are given in
Figure~\ref{fig:ECEC}. This factorization reduces the memory consumption
as both structured sparse matrix $C$ and the structured dense matrix
$E^c$ can be efficiently represented.
\arxiv{
Furthermore, if the matrix $E^c$
is block-diagonal and $C$ is unweighted, the structure of $C$ allows embedding lookup with
no additional computation cost. We fix the locations of non-zero entries in $C$ and do not
change them during training. We train
$E^c$ and values of non-zero entries in $C$ using stochastic gradient descent.
Next, we describe the
structure of the sparse and dense matrices.}
\begin{figure}[t]
  \begin{eqnarray}
    \arxiv{
\underbrace{\begin{blockarray}{cccc}
\begin{block}{(cccc)}
\textcolor{red}{0.1}   & \textcolor{red}{1.5}  & \textcolor{green}{1.0}  & \textcolor{green}{-3.2}\\
  \textcolor{blue}{-1.8} & \textcolor{blue}{2.0} & \textcolor{blue}{-1.8} & \textcolor{blue}{2.0}   \\
  \textcolor{green}{1.0}  & \textcolor{green}{-3.2} & \textcolor{red}{0.1}   & \textcolor{red}{1.5} \\
  \textcolor{red}{\bf 0.1}   & \textcolor{red}{\bf 1.5} & \textcolor{blue}{\bf -1.8} & \textcolor{blue}{\bf 2.0}  \\
  \textcolor{red}{0.1}   & \textcolor{red}{1.5} & \textcolor{red}{0.1}   & \textcolor{red}{1.5} \\
  \textcolor{blue}{-1.8} & \textcolor{blue}{2.0} & \textcolor{green}{1.0}  & \textcolor{green}{-3.2} \\
\end{block}
\end{blockarray}}_E
=
}
\underbrace{
\begin{blockarray}{ccc|ccc}
\begin{block}{(ccc|ccc)}
  \textcolor{red}{1} & \textcolor{red}{0} & \textcolor{red}{0} & \textcolor{green}{0} & \textcolor{green}{1} & \textcolor{green}{0}  \\
  \textcolor{blue}{0} & \textcolor{blue}{0} & \textcolor{blue}{1} & \textcolor{blue}{0} & \textcolor{blue}{0} & \textcolor{blue}{1}  \\
  \textcolor{green}{0} & \textcolor{green}{1} & \textcolor{green}{0} & \textcolor{red}{1} & \textcolor{red}{0} & \textcolor{red}{0}  \\
  \textcolor{red}{\bf 1} & \textcolor{red}{\bf 0} & \textcolor{red}{\bf 0} & \textcolor{blue}{\bf 0} & \textcolor{blue}{\bf 0} & \textcolor{blue}{\bf 1} \\
  \textcolor{red}{1} & \textcolor{red}{0} & \textcolor{red}{0} & \textcolor{red}{1} & \textcolor{red}{0} & \textcolor{red}{0}  \\
  \textcolor{blue}{0} & \textcolor{blue}{0} & \textcolor{blue}{1} & \textcolor{green}{0} & \textcolor{green}{1} & \textcolor{green}{0} \\
\end{block}
\end{blockarray}
}_{C(k=3, n=2)}
\arxiv{\hspace{0.5cm}}
\underbrace{
\begin{blockarray}{cc|cc}
\begin{block}{(cc|cc)}
    \textcolor{red}{0.1}  & \textcolor{red}{1.5}  & \BAmulticolumn{2}{c}{\multirow{3}{*}{$\bf 0$}} \\
    \textcolor{green}{1.0}  & \textcolor{green}{-3.2} & & \\
    \textcolor{blue}{-1.8} & \textcolor{blue}{2.0}  & & \\
    \cline{1-4}
    \BAmulticolumn{2}{c|}{\multirow{3}{*}{$\bf 0$}}   & \textcolor{red}{0.1}  & \textcolor{red}{1.5} \\
    & & \textcolor{green}{1.0}  & \textcolor{green}{-3.2}\\
    & & \textcolor{blue}{-1.8} & \textcolor{blue}{2.0}\\
    \end{block}
\end{blockarray}
}_{E^{c}}
\nonumber
\end{eqnarray}
\caption{An example of WEST factorization when $V= 6$, $d= 4$, $k = 3$, and $n = 2$.}
\label{fig:ECEC}
\end{figure}
\conference{\vspace{-0.3cm}}
\subsection{Structured sparse matrix}
\label{sec:codebook}
Let $c: [1,V] \to \cup_{i\leq n} [1,k]^i$ be a mapping from
  the vocabulary to the set of sequences of length at most $n$, where
  each entry ranges from $1$ to $k$. We refer to $n$ as the
  \emph{code length} and $k$ as the \emph{alphabet size}. Let $c_i(w)$ be
  the $\text{i}^\text{th}$ entry of the code for word $w$
  \footnote{In the rest of the paper, we use "word" for categorical
  features and output classes as our main application is language models.}
  . Given such a
  codebook $c$, we construct a sparse matrix $C$ of size $V \times n
  \cdot k$ as follows. For $i \leq n$ and $j \leq k$,
  \[
C_{w, (i-1)\cdot k+j} \neq 0 \text{ if and only if } c_{i}(w) = j.
\]
Furthermore, we define $\lambda_{w, i}$ to be the weight corresponding
to the entry corresponding to $c_{i}(w)$.
We differentiate between two types of sparse code books, the \textit{weighted}
sparse matrix where the non-zero entries can take any value and the
\textit{unweighted} sparse matrix where the non-zero entries are restricted to
be
one, i.e., $\lambda_{w, i}=1 \,\, \forall w, \, \forall i$. \arxiv{Figure~\ref{fig:sparse}}\conference{Figure~\ref{fig:ECEC}}
has an example of a unweighted structured sparse matrix $C$\arxiv{ and the
corresponding codebook $c$}.

To store the codebook (the sparse matrix entries indices), \arxiv{\[}\conference{$} V
\cdot n \cdot \lceil \log_2 k \rceil \arxiv{\]}\conference{$}
bits are needed. This can also be prohibitive in many applications. Hence, we
propose to use sparse codebooks that can be stored succinctly with
fewer than $V \cdot n$ additional parameters e.g., language codes and
random codes.

\arxiv{
\begin{figure}[h]
  \begin{eqnarray}
      \underbrace{
\begin{blockarray}{ccc}
\arxiv{}  & c_1 \left( w \right) & c_2 \left( w \right) \\
\arxiv{\begin{block}{c(cc)}}
\conference{\begin{block}{ccc}}
\arxiv{  \texttt{I}   \,\,\,}&  1 & 2 \\
\arxiv{  \texttt{It}  \,\,\,} & 3 & 3 \\
\arxiv{  \texttt{He}  \,\,\,}&  2 & 1 \\
\arxiv{  \texttt{She} \,\,\,}& {1} & {3} \\
\arxiv{  \texttt{You} \,\,\, }& 1 & 1 \\
\arxiv{  \texttt{They}\,\,\,} & 3 & 2 \\
\end{block}
\end{blockarray}
}_{c(3,2)}
      =
\underbrace{
      \begin{blockarray}{ccc|ccc}
 & {{\mathbbm{1}}_{c_1 \left( w \right)}} & & & {{\mathbbm{1}}_{c_2 \left( w \right)}} & \\
\begin{block}{(ccc|ccc)}
   \textcolor{red}{1} & \textcolor{red}{0} & \textcolor{red}{0} & \textcolor{green}{0} & \textcolor{green}{1} & \textcolor{green}{0}  \\
   \textcolor{blue}{0} & \textcolor{blue}{0} & \textcolor{blue}{1} & \textcolor{blue}{0} & \textcolor{blue}{0} & \textcolor{blue}{1}  \\
\textcolor{green}{0} & \textcolor{green}{1} & \textcolor{green}{0} & \textcolor{red}{1} & \textcolor{red}{0} & \textcolor{red}{0}  \\
 \textcolor{red}{\bf 1} & \textcolor{red}{\bf 0} & \textcolor{red}{\bf 0} & \textcolor{blue}{\bf 0} & \textcolor{blue}{\bf 0} & \textcolor{blue}{\bf 1} \\
 \textcolor{red}{1} & \textcolor{red}{0} & \textcolor{red}{0} & \textcolor{red}{1} & \textcolor{red}{0} & \textcolor{red}{0}  \\
\textcolor{blue}{0} & \textcolor{blue}{0} & \textcolor{blue}{1} & \textcolor{green}{0} & \textcolor{green}{1} & \textcolor{green}{0} \\
\end{block}
      \end{blockarray}
}_{C(3,2)}
\nonumber
  \end{eqnarray}
  \conference{\vspace{-0.2cm}}
  \caption{An example codebook and the corresponding sparse structured
    matrix for $V=6$, $n=2$, and $k=3$.}
    \label{fig:sparse}
\end{figure}
}

\noindent {\bf Language coding:} Words are composed of characters and in many applications it is
sufficient to use characters in the word itself for embedding.  To
be more specific, let $F$ be a collection of sub-units such as
characters or word-pieces~\cite{schuster2012japanese} that can be concatenated to form words.  We
then decompose a word into sub-units such as characters or word-pieces
and obtain the code by mapping each word into sub-units. More formally
let $w=w_1,w_2,\ldots w_{n'}$ where each $w_i \in F$ and $n'
\leq n$. Let $F(w_i)$ be the index of $w_i$ in $F$, then
\[
c(w) = F(w_1), F(w_2),\ldots, F(w_{n'}).
\]
For example, if the set of words are $\{$i, it, he, she, you, they$\}$,
and sub-units are $\{$I, t, he, s, you, y$\}$, then $c(\text{she})$ is
$(4,3)$. The advantage of the language codebook is that the word-to-character mapping
is typically stored along with the network and hence
does not require any additional space.

\noindent {\bf Random coding:} In many applications such as language models, there are many layers
after the embedding layer. In these applications, even if the sparse
matrix constrains the modeling capacity, the following sequence of
non-linear layers compensate for any insufficiency.
Following~\cite{li2017slim}, we employ random
codes where each word is mapped to a random sequence of length
$n$ such that no two words have the same code.\arxiv{\newline
\begin{center}\fbox{\parbox{0.9\textwidth}{
\vspace{-1ex}
\begin{center}
Rand($k$,$n$)
\end{center}
\vspace{-2ex}
For each word
      $w$:
\begin{enumerate}
\item For each $ i \leq n$, draw a uniform sample from $[1,k]$ and
  denote it as $c_i$.
\item Compute the concatenated codeword ${\bf c} \tdef \left(
c_1,c_2\ldots c_n\right)$.
\item If ${\bf c}$ is already assigned to another word, go back to
step (1), else assign $C\left(w\right) = {\bf c}$.
\end{enumerate}
  }}\end{center} \vspace{2ex} \par
} The resulting codebook is uniquely decodable by design\arxiv{ since
each embedding input is represented by a unique combination of
parameters that are not identical to any other embedding input}.
Furthermore, for any two words $w_1, w_2$, the probability of collision at any index is
\conference{:$}\arxiv{\[}\text{Prob} \left( c_i ( w_1 ) = c_i ( w_2 ) \right) = 1 / k,\conference{$}\arxiv{\]} for $i \leq n$. Note
that for alphabet size $k$, this is the minimum possibility of
collision for any algorithm.  The advantage of the random codebook is
that it can be generated on-the-fly, given a seed. Hence, storing random
codebook needs just $V$ parameters.

\conference{\vspace{-0.25cm}}
\subsection{Structured dense matrix}

We propose to use structured dense matrices $E^c$ that can be
succinctly represented. We focus on two examples.

\noindent\textbf{ Block-diagonal structure~\cite{li2017slim}}: Let
$E^i$s be the \emph{sub-unit embedding matrices} of size $k \times d/n$,
then $E^c$ is a block-diagonal matrix with $E^i$s as the diagonal
entries i.e., \[ E^c_{(i-1)\cdot k + i', (i-1)\cdot d/n + j'} = E^{i}_{i', j'},
\]
for $i \leq n$, $i' \leq k$ and $j' \leq d/n$, else $0$. If $c(w) = c_1(w),\ldots,c_n(w)$ is the code for $w$ and $\lambda_{w,i}$
is the corresponding weights in the structured sparse matrix,
then the embedding for word $w$ with block diagonal matrix is
\begin{eqnarray}
\label{eq:con}
E_w = \left[ \lambda_{w,1} E^1_{c_1(w)}; ...; \lambda_{w,n} E^n_{c_n(w)} \right],
\end{eqnarray}
which is weighted concatenation of the rows of the sub-unit embedding matrices corresponding
to the code $c(w)$. Note that the concatenation using unweighted sparse matrix
does not require any extra computation.  \arxiv{Figure~\ref{fig:block} shows an example of the
 block-diagonal structure.}

\noindent \textbf{Band structure: } Let $E^i$s be matrices of size $k \times d$, then
  $E^c$ is the matrix obtained by stacking entries of $E^i$s one below
another i.e., \[ E^c_{(i-1)\cdot k + i', j} = E^i_{i', j}.
\]
Under the band structure, the embedding can be computed as
\begin{eqnarray}
\label{eq:sum}
E_w = \lambda_{w,1} E^1_{c_1(w)} +  ... + \lambda_{w,n} E^n_{c_n(w)},
\end{eqnarray}
which is the weighted sum of rows
corresponding to the code $c(w)$.
Note that to store the structured dense matrix $E^c$ it is sufficient
to store the sub-unit embedding matrices $E^i$s and hence the space used
in the block-diagonal structure is \arxiv{\[}\conference{$} k \times
d \arxiv{,\]}\conference{$} and the space used by the band structure
is \arxiv{\[}\conference{$} k \times d \times n.
\arxiv{\]}\conference{$} \arxiv{Furthermore, if}\conference{If} the parameters are tied i.e.,
all $E^i$s are equal, then the number of parameters reduces by a
factor of $n$. \arxiv{Figure~\ref{fig:band} shows an example of the
 band structure.}

\arxiv{
\begin{figure}[t]
  \centering
  \subfigure[Block-diagonal structure]{
\begin{blockarray}{ccccccccc}
    \begin{block}{c(cccccccc)}
      &  \BAmulticolumn{2}{c}{\multirow{1}{*}{$ E^1$}} & \BAmulticolumn{2}{c}{\multirow{1}{*}{$ 0$}} & \BAmulticolumn{2}{c}{\multirow{1}{*}{\dots}} & \BAmulticolumn{2}{c}{\multirow{1}{*}{$ 0$}} \\
      & \BAmulticolumn{2}{c}{\multirow{1}{*}{$ 0$}}   & \BAmulticolumn{2}{c}{\multirow{1}{*}{$ E^2$}} & \BAmulticolumn{2}{c}{\multirow{1}{*}{\dots}}& \BAmulticolumn{2}{c}{\multirow{1}{*}{$ 0$}}\\
      & \BAmulticolumn{2}{c}{\multirow{1}{*}{\vdots}}    & & \BAmulticolumn{2}{c}{\multirow{1}{*}{$\ddots$}} & & \BAmulticolumn{2}{c}{\multirow{1}{*}{\vdots}} \\
      & & & & \\
      & \BAmulticolumn{2}{c}{\multirow{1}{*}{$ 0$}}   & \BAmulticolumn{2}{c}{\multirow{1}{*}{$ 0$}}  & \BAmulticolumn{2}{c}{\multirow{1}{*}{\dots}} & \BAmulticolumn{2}{c}{\multirow{1}{*}{$ E^{n}$}}\\
    \end{block}
\end{blockarray}
\label{fig:block}
}
\hspace{3.0cm}
\subfigure[Band structure]{
    \begin{blockarray}{cccccc}
    \begin{block}{(cccccc)}
        \BAmulticolumn{2}{c}{\multirow{1}{*}{\dots}} & \BAmulticolumn{2}{c}{\multirow{1}{*}{$ E^1$}} & \BAmulticolumn{2}{c}{\multirow{1}{*}{\dots}} \\
        \BAmulticolumn{2}{c}{\multirow{1}{*}{\dots}} & \BAmulticolumn{2}{c}{\multirow{1}{*}{$ E^2$}} & \BAmulticolumn{2}{c}{\multirow{1}{*}{\dots}} \\
       \BAmulticolumn{2}{c}{\multirow{1}{*}{\dots}} & \BAmulticolumn{2}{c}{\multirow{1}{*}{\vdots}} & \BAmulticolumn{2}{c}{\multirow{1}{*}{\dots}} \\
      & & & & \\
        \BAmulticolumn{2}{c}{\multirow{1}{*}{\dots}} & \BAmulticolumn{2}{c}{\multirow{1}{*}{$ E^{n}$}} & \BAmulticolumn{2}{c}{\multirow{1}{*}{\dots}} \\
    \end{block}
    \end{blockarray}
  \label{fig:band}
}
  \caption{Examples of structured dense matrices}
\end{figure}
}

\subsection{WEST interpretation}
\label{sec:softmax}
As stated in the previous section, WEST for embedding layer can be
interpretted as encoding the categorical features with sub-units or
codes. The embedding of the categorical feature can be obtained either
by concatenation~\eqref{eq:con} or sum of sub-unit embeddings~\eqref{eq:sum}.

The same applies for classification weights of the softmax layer. Further, WEST for softmax layer
can be interpreted as
MaxEnt model over sub-unit weights. This distinguishes
 WEST from character or word models in the way it models the
probability of next word given the history. If $h$
is the penultimate layer activation and $E^c$ is a band structured matrix, the logit for
class $w$ is
\begin{eqnarray*}
l_w = E_w \cdot h  = \sum_{i=1}^{n} \lambda_{w, i} E^i_{c_i(w)} \cdot h,
\end{eqnarray*}
and the posterior probability is
\begin{eqnarray}
  P (w | h) &=& \frac{\exp (l_w)} {\sum_{w'} \exp(l_{w'})} \nonumber \\
  &=& \frac{1}{\mathbbm{Z}(h)} \exp\left(\sum^n_{i=1} \lambda_{w, i} E^i_{c_i(w)} \cdot h \right),
\label{eq:exp}
\end{eqnarray}
where $\mathbbm{Z}(h)$ is the normalization factor.

We illustrate the difference between WEST and other approaches with a simple example where
the goal is to predict the word \texttt{west} given a history $h$.
One way is doing it directly,
by assigning one embedding vector per word. This may not be the best approach, due to large memory
footprints and data sparsity. The alternative is using sub-units like characters and estimate
this probability $P (\texttt{west} | h)$ via chain rule:
\begin{eqnarray*}
\label{eq:cat}
  P(\texttt{w} | h) \cdot P (\texttt{e} | h, \texttt{w}) \cdot P (\texttt{s} | h, \texttt{w}, \texttt{e})
  \cdot P (\texttt{t} | h, \texttt{w}, \texttt{e}, \texttt{s})
  \cdot P (\texttt{<eow>} | h, \texttt{w}, \texttt{e}, \texttt{s},\texttt{t}  ),
\end{eqnarray*}
where \texttt{<eow>} is the end of word symbol.
Estimating the word probability with the above equation is difficult as the recurrent model needs to
remember longer contexts of sub-units. Furthermore, since these models are normalized on sub-units, they
assign probabilities to character sequences that are not
in the vocabulary e.g., "\texttt{wese}". Similar to character or word-piece models, WEST uses sub-units,
but the predictions are over larger units (e.g, words). This has two advantages:
it has a smaller memory footprint as it uses
sub-units internally and it has modelling advantage as it outputs normalized probabilities over larger units.

In WEST, $P(\texttt{west} | h)$ is estimated via Eq.~\eqref{eq:exp}.
This is very similar to MaxEnt framework~\cite{berger1996maximum} where the MaxEnt
features are the logits assigned to the sub-units. Initial network components such as
embedding and recurrent units act as feature extractors and compute logits
per sub-unit. The final layer acts as a MaxEnt model over sub-units, with MaxEnt
features being the logits assigned to the sub-units by the initial network components.
Both the feature extractors and the MaxEnt model are trained jointly. A natural question is
how much the choice of sub-units matter? We explore this in the next section.

\section{Experiments}
\label{sec:experiments}
Experiments are designed to evaluate different aspects of WEST compression
in terms of \textit{maximum achievable compression rate} for embedding
and softmax, \textit{effect of word-level normalization},
\textit{choice of sub-units} and
performance for an \textit{embedded ASR task}.

\noindent {\bf Datasets.}
We use {\textit{Penn Treebank} (PTB)} public dataset that\footnote{Downloaded from:
  \url{http://www.fit.vutbr.cz/~imikolov/rnnlm/simple-examples.tgz}}
  \cite{marcus1993building}
consists of $929$k words in the training corpus and $82$k words in the test corpus
with $10$k vocabulary size. The model from~\cite{zaremba2014recurrent} is used as
the baseline (see Table 1 in~\cite{zaremba2014recurrent} for
details). The {\textit{embedded task}} presented here is a state-of-the-art ASR on the phone which uses
a RNN LM for second pass rescoring. The vocabulary size is $64$k and the model size
is $15$ MB on disk (Embedding: $4.1$ MB, LSTMs: $8.45$ MB and Softmax: $4.1$ MB).
\arxiv{The embedding
matrix has dimensions $64\text{k} \times 16$ and the classification matrix has dimensions $64\text{k} \times 32$.
The recurrent architecture is a stack of three layers of LSTMs with $512$ cells
per layer and $32$ dimensional projection layer between LSTM layers.}
The WER for this model is reported on a set of $14$k anonymized, hand-transcribed
voice search utterances extracted from live traffic \cite{mcgraw2016personalized}.
\arxiv{The WER without rescoring on this task
is $16.4 \%$ which drops to $15.1 \%$ after rescoring with the above described RNN LM.}

\begin{figure}[t]
  \centering
  \subfigure[Compression for different code-lengths]{
    \includegraphics[width=0.4\linewidth]{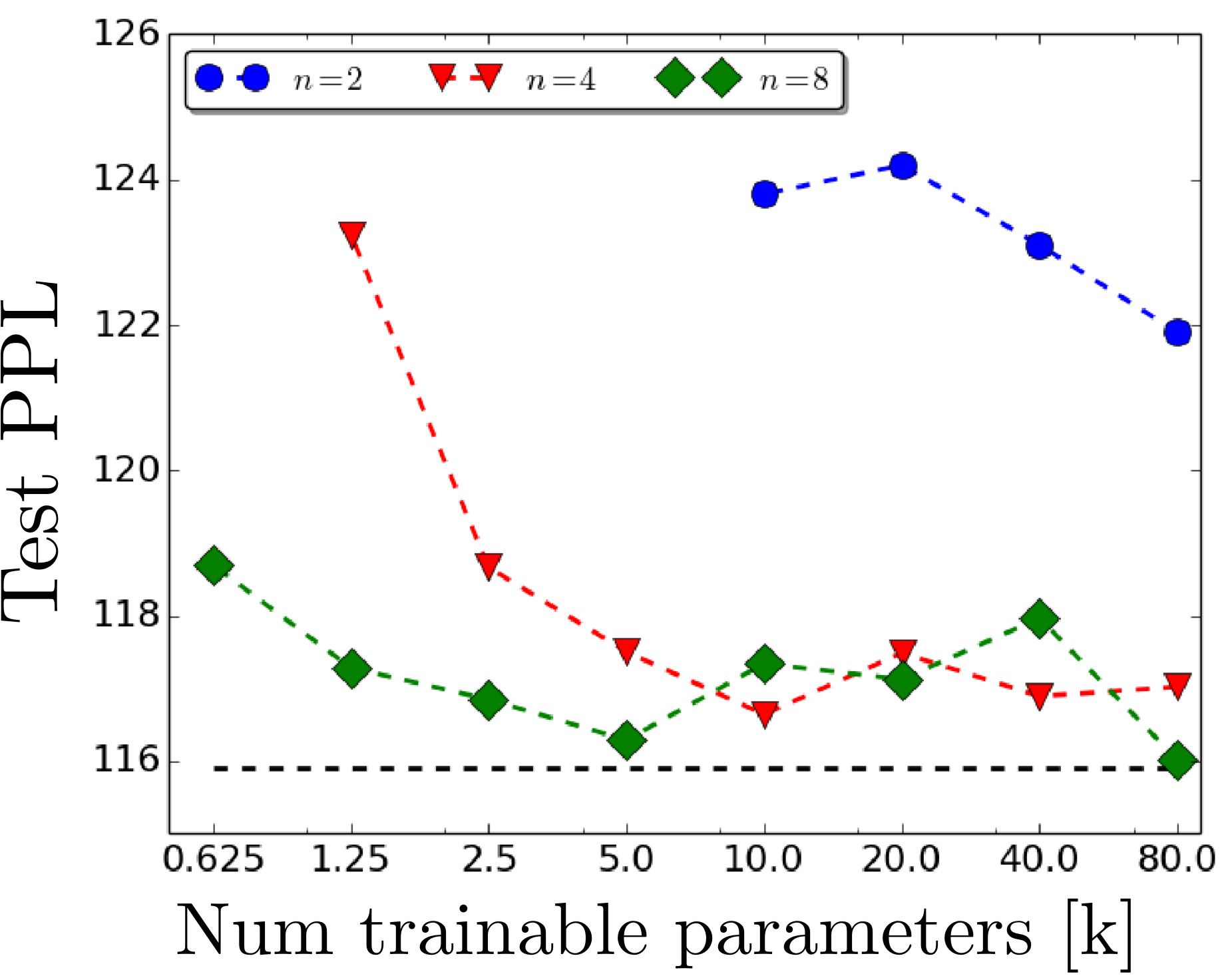}
    \label{fig:ptb_embedding_compression}
  }
  \subfigure[Generalization ($n = 8$)]{
    \includegraphics[width=0.4\linewidth]{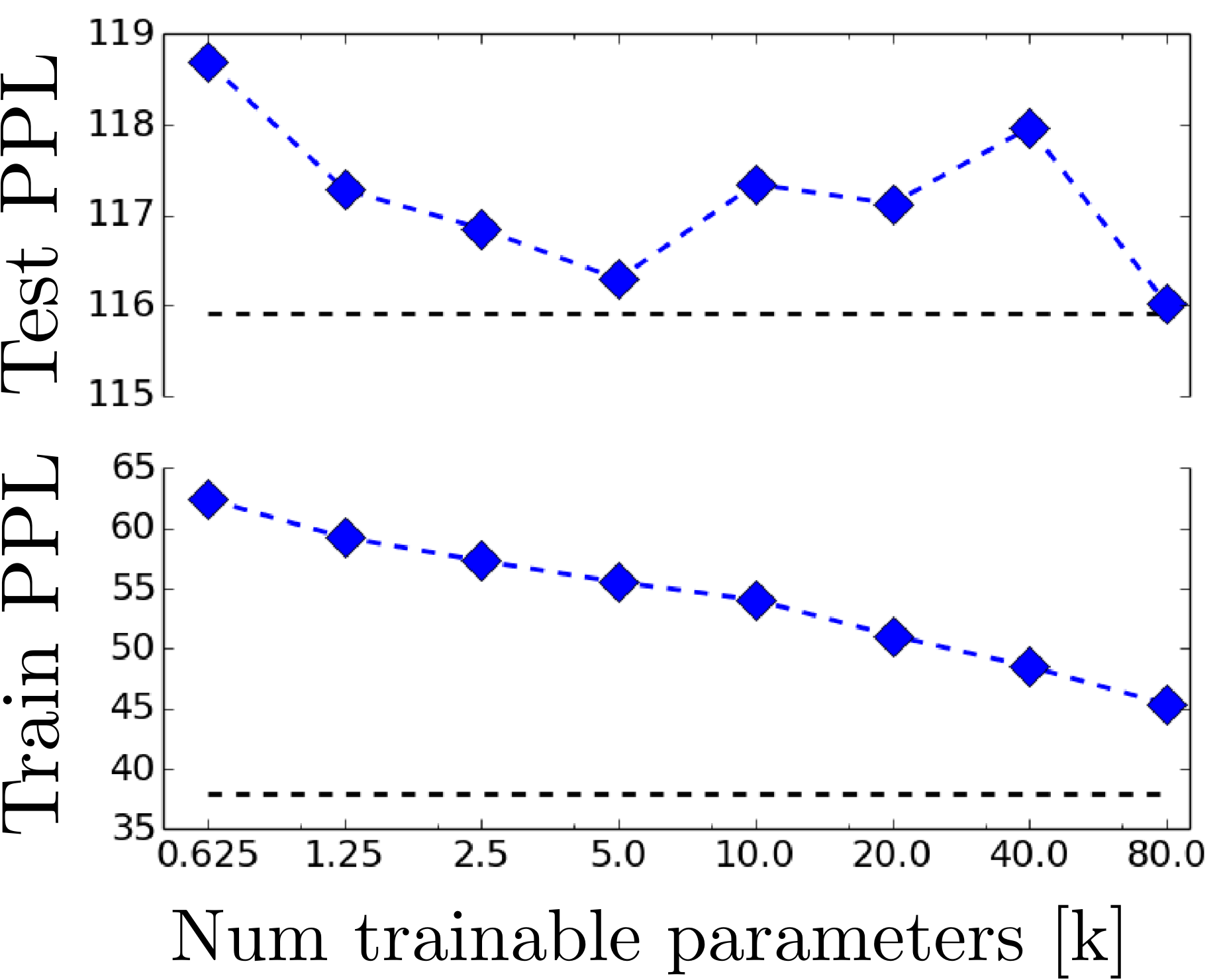}
    \label{fig:ptb_embedding_generalization}
  }
  \caption{WEST embedding performance on PTB with random codes.
  WEST models use an additional $10$k non-trainable parameters
  to store the seeds for random codes.}
  \label{fig:ptb_embedding}
\end{figure}

\noindent {\bf Maximum compression rate:}
The performance of WEST embedding with random codes, unweighted sparse matrices, and tied block-diagonal structure for
PTB is presented in Figure ~\ref{fig:ptb_embedding}.
The number of trainable parameters in the embedding matrix can be compressed up to $1000$ times with perplexity (PPL)
close to baseline (dashed horizontal line). At the same compression ratio,
longer code length achieves better perplexity
as larger $n$ increases sampling space size $k ^ n$, and thus reduces the collision
probability. Comparing the gap between the train and test perplexity for WEST models with baseline suggests
that the baseline model is prone to over-fitting, while WEST models achieve similar test
PPL with larger training PPL and hence generalize better, see Figure~\ref{fig:ptb_embedding_generalization}.
For softmax, the maximum compression rate which achieves the same performance as
baseline is about two (fourth row of Table~\ref{tab:sub-units}), where
the most frequent $2000$ words are coded as themselves and rest are coded using random codes.

\noindent {\bf Effect of word-level normalization:}
To examine the effect of word-level normalization \eqref{eq:exp},
we design the following experiment.
For each word $w$, we first optimize the model that estimates $P(w | h)$
by estimating the character probability individually i.e., for the word \texttt{west}, we estimate probability as
\begin{eqnarray*}
\label{eq:cat}
  P_1(\texttt{w} | h) \cdot P_2 (\texttt{e} | h) \cdot  P_3 (\texttt{s} | h) \cdot   P_4 (\texttt{t} | h) \cdot
  P_5 (\texttt{<eow>} | h),
\end{eqnarray*}
where $P_i(c|h)$ is the probability of $i^\text{th}$ character being $c$ with history $h$.
This corresponds to the band structure without normalization. The word level test PPL
of this model is $366$k (char-normalized in Table~\ref{tab:west_softmax}).
However, if we train the model using
WEST~\eqref{eq:exp}, the performance improves to $533$. We further added word-level
bias term and sparse weights $\lambda_{w, i}$. This increases the number of parameters only by $0.07$M,
but improves the perplexity to $149.72$. By increasing the number of parameters of LSTM,
the perplexity can be further improved to $114.59$ while keeping the model size same
as the baseline.
\begin{table}[ht]
  \centering
  \begin{tabular}{|c|c|c|c|c|c|}
    \hline
    Model              &   Test PPL & \multicolumn{3}{|c|}{\# trainable params [M]}  \\
    \cline{3-5}
                       &            &  Emb  &  LSTM  &  Soft           \\
    \hline
    Char-normalized   &    366k    &  2.0  &   0.5  &  0.20  \\
    \hline
    Word-normalized    &    533.39  &  2.0  &   0.5  &  0.20  \\
    + Word biases      &    238.06  &  2.0  &   0.5  &  0.21  \\
    + Sparse weights   &    149.72  &  2.0  &   0.5  &  0.27 \\
    + Wider lstm       &    114.59  &  2.0  &   2.24 &  0.27  \\
    \hline
    Baseline           &    115.91  &  2.0  &   0.5  &  2.01  \\
\hline
\end{tabular}
\caption{Softmax with language codes.}
\label{tab:west_softmax}
\end{table}

\noindent{\bf Choice of sub-units.}
Table~\ref{tab:sub-units} presents the effect of various coding schemes. The first row
is the performance of the character codebook with bias and weights as in Table~\ref{tab:west_softmax}.
The second row uses random codes with alphabet size $49$ (which is equal to number
of unique characters in PTB) and code length of $12$ which matches the total
parameters of the character coding.
The PPL for the random coding is better than
the character model (row 2 and row 1 in Table~\ref{tab:sub-units}).
This shows that the model need not use the language structure
and it can learn features with random encoding. Let Rand($k$, $n$, $t$), denote the
code where the most frequent $t$ words are coded as themselves and rest of the words are assigned randomly.
To be more concrete, let words be ordered in decreasing order for frequency. For $w \leq t$, let $c(w) = (k + w)$
and for $w > t$, the codes are randomly assigned using {Rand}$(k,n)$. Note that under this construction
the alphabets assigned for top $t$ words are unique and are not shared with any other codes.
As shown in
row 3 and 4 in Table~\ref{tab:sub-units}, the PPL gets better.
Finally compressing both embedding and softmax layers
and distributing the saved parameters to the LSTM, the PPL reduces to $92.36$,
which is the best test PPL for this model size for PTB to the best of our knowledge.
\begin{table}[ht]
\small
  \centering
  \begin{tabular}{|c|c|c|c|c|c|}
    \hline
    Sub-unit              &   Test PPL & \multicolumn{3}{|c|}{\# trainable params [M]}  \\
    \cline{3-5}
                                 &            &  Emb  &  LSTM  &  Soft           \\
    \hline
    Character                    &    149.72  &  2.0  &   0.5  &  0.27 \\
    Rand($49$, $12$)             &    134.58  &  2.0  &   0.5  &  0.25  \\
    \hline
    Rand($49$, $12$, $2000$)  &    121.73  &  2.0  &   0.5  &  0.63 \\
    Rand($49$, $12$, $4000$)  &    116.84  &  2.0  &   0.5  &  1.00 \\
    \hline
    Rand($49$, $12$)             &    101.58  &  2.0  &   2.24 &  0.25 \\
    Rand($49$, $12$, $2000$)  &    92.36   &  0.5  &   3.38 &  0.63 \\
    \hline
    Baseline         &    115.91  &  2.0  &   0.5  &  2.01  \\
\hline
\end{tabular}
\caption{Performance of different sub-units, character, Rand$(k, n)$ and
Rand$(k, n, t)$. All models use the band structure for the softmax layer.}
\label{tab:sub-units}
\conference{\vspace{-0.5cm}}
\end{table}
\begin{table}[ht]
  \centering
  \hspace{0.5cm}
  \begin{tabular}{|c|c|c|c|c|c|c|}
    \hline
    Structure             &  {PPL} & \multicolumn{2}{|c|}{WER [\%]} & Size\\
    \cline{3-4}
                          &  Test        &    VS     &   Dict & [MB]\\
    \cline{1-5}
    Baseline              &  68.07       &   13.7    &  7.3 & 15 \\
    \hline
    + Embedding($15.5X$)   &  70.1        &   13.7    &  7.3 & 11.25 \\
    + Softmax($3.1X$)       &  71.0        &   13.6    &  7.3 & 4.75 \\
    + Quantization        &  71.0        &   13.7    &  7.3 & 1.35 \\
    \hline
    Reallocation   &  56.3        &  13.3     &  7.1 & 15 \\
\hline
\end{tabular}
\caption{WEST for on-device second pass rescoring.}
\label{tab:ondevice}
\end{table}

\noindent{\bf Large-scale embedded ASR.}
Table~\ref{tab:ondevice} presents performance of WEST compression
on a second pass RNN LM of a large-scale on-device ASR task. The number of trainable
parameters in the embedding layer is compressed $500$ times using WEST with random codes,
unweighted sparse matrices,
and tied block diagonal structure.
This doesn't change the WER and saves $4$ MB.
Storing the random codebook needs another $64$k parameters.
This gives a compression of $15.5$ times for the embedding layer.
For softmax, we use band structure and language codes where the sub-units
include most frequent $16$k words
and characters. This results in a
moderate softmax compression of $3.1$ and the model size
drops to $4.75$ MB. Note that we do not need extra space for storing the language codebook
as the information is already present in the symbol table.
Finally, to demonstrate the flexibility of WEST to
integrate with other compression techniques, we apply scalar quantization
on top of already compressed WEST model to reduce the model size to $1.35$ MB,
resulting in a total of $11$ times compression without any performance degradation. If the
parameter savings from embedding and softmax is used in LSTMs while keeping the
same size of the baseline model, WER improves by $3 \%$ relative (last row of Table~\ref{tab:ondevice}).
We remark that we obtained similar compression rates using random codes instead of language codes.

\conference{\vspace{-1ex}}
\section{Conclusion}
\conference{\vspace{-1ex}}
We proposed WEST, a single-stage compression technique for reducing the size of
embedding and softmax
layers for large vocabulary tasks. WEST bridges the gap between larger unit and
smaller sub-unit models and provides a general framework that can be interpreted
as MaxEnt model over sub-units.
We reported significant compression rates for
embedding and moderate rates
for the softmax layer. Our experiments showed that the choice of sparse and dense matrices
do not matter for the embedding layer, but they do matter for the softmax layer.
In particular for softmax compression,
variations of random codes performed slightly better than character codes and
weighted sparse matrices performed significantly
better than unweighted sparse matrices.
Finally, we demonstrated that beside compression, while keeping overall model
size fixed, reallocating saved parameters from WEST
to other network components can result in even better performance.

\section{Acknowledgements}

Authors would like to thank Jayadev Acharya, Michieal Bacchiani, Tom Bagby,
 Ciprian Chelba, Badih Ghazi,
 Yanzhang He, Shankar Kumar, Rina Panigrahy, Yuan Shangguan, Matt Shannon,
 and Ke Wu
for helpful discussions and comments.

\bibliographystyle{IEEEbib}
\bibliography{west}

\end{document}